%% file: ijcnn.tex
\documentclass[conference]{IEEEtran}
\IEEEoverridecommandlockouts
\usepackage{cite}
\usepackage{amsmath,amssymb,amsfonts}
\usepackage{algorithmic}
\usepackage{graphicx}
\usepackage{textcomp}
\usepackage{xcolor}
\def\BibTeX{{\rm B\kern-.05em{\sc i\kern-.025em b}\kern-.08em
    T\kern-.1667em\lower.7ex\hbox{E}\kern-.125emX}}
\IEEEoverridecommandlockouts
\usepackage[numbers]{natbib}
\usepackage{amsmath,amssymb,amsfonts}
\usepackage{algorithmic}
\usepackage{graphicx}
\usepackage{textcomp}
\usepackage{xcolor}
\usepackage{multirow}
\usepackage{tikz}
\usepackage{pgfplots}
\usepackage{amssymb}
\usepackage{subfigure}
\begin{document}

\title{Intra-Model Collaborative Learning of Neural Networks}
\author{
    \IEEEauthorblockN{Shijie Fang\IEEEauthorrefmark{2}, Tong Lin\IEEEauthorrefmark{2}\IEEEauthorrefmark{1}}
    \IEEEauthorblockA{\IEEEauthorrefmark{2}Key Laboratory of Machine Perception (Ministry of Education), Beijing, China} 
    \IEEEauthorblockA{\IEEEauthorrefmark{1}Peng Cheng Laboratory, Shenzhen, China}
    \IEEEauthorblockA{fangshijie@stu.pku.edu.cn, lintong@pku.edu.cn}
}

\maketitle
\input{sections/abstract.tex}
\input{sections/introduction.tex}
\input{sections/prior_work.tex}
\input{sections/method.tex}
\input{sections/experiments.tex}
\input{sections/conclusion.tex}

\section*{Acknowledgment}
This work was supported by NSFC Tianyuan Fund for Mathematics (No. 12026606), and National Key R\&D Program of China (No. 2018AAA0100300).

\bibliographystyle{IEEEtran}
\bibliography{ijcnn}

\end{document}

%% file: sections/abstract.tex
\begin{abstract}
  Recently, collaborative learning proposed by Song and Chai has achieved remarkable improvements in image classification tasks by simultaneously training multiple classifier heads. However, huge memory footprints required by such multi-head structures may hinder the training of large-capacity baseline models. The natural question is how to achieve collaborative learning within a single network without duplicating any modules. In this paper, we propose four ways of collaborative learning among different parts of a single network with negligible engineering efforts. To improve the robustness of the network, we leverage the consistency of the output layer and intermediate layers for training under the collaborative learning framework. Besides, the similarity of intermediate representation and convolution kernel is also introduced to reduce the reduce redundant in a neural network. Compared to the method of Song and Chai, our framework further considers the collaboration inside a single model and takes smaller overhead. Extensive experiments on Cifar-10, Cifar-100,  ImageNet32 and STL-10 corroborate the effectiveness of these four ways separately while combining them leads to further improvements. In particular, test errors on the STL-10 dataset are decreased by $9.28\%$ and $5.45\%$ for ResNet-18 and VGG-16 respectively. Moreover, our method is proven to be robust to label noise with experiments on Cifar-10 dataset. For example, our method has $3.53\%$ higher performance under $50\%$ noise ratio setting.
\end{abstract}

\begin{IEEEkeywords}
  collaborative learning, neural networks, machine learning
\end{IEEEkeywords}

%% file: sections/introduction.tex
\section{Introduction}
Despite the great success of deep neural networks, their training remains to be difficult 
both practically and theoretically. It is well-known that ensembling neural networks \cite{ensemble_learning}
or enlarging the capacity of networks \cite{wide_res_net, resnext} tends to yield better performance. 
However, these methods lead to a relatively expensive extra computation cost in both training 
and test, which prevents them from deploying in real settings.
It is challenging how to achieve improvements without any extra 
cost in inference with the capacity and computation of the network kept unchanged.

To address this challenge, Song and Chai \cite{collaborative} proposed \textbf{collaborative learning}, where 
multiple hierarchical subnets of the network are simultaneously trained to improve the generalization. 
Using such a collaborative learning framework, they achieved $26.36\%$ test error on CIFAR-100 with ResNet-32. 
Despite the great achievement, memory and computation cost for training such a multi-instance or tree structure
with multiple paths is too expensive when the number of subnets grows. 
For example, training a network with four paths will take approximately 1.5 times of memory beyond the baseline, 
which is prohibitive for training a large baseline model on GPUs with limited memory.

The natural question is how to perform collaborative learning with little extra memory cost in training. 
In this work, we proposed four effective ways of collaborative learning in a single network. 
To be more specific, we consider the collaboration of output layer, intermediate layers, feature representations, and convolution kernel:
\begin{itemize}
  \item For the collaboration of the output layer, we use the simple yet effective dropout technique to replace the tree-structure used in the method of Song and Chai for saving expenses in training and inference. The consistency of multiple inputs generated by dropout is regularized to produce similar output to improve the robustness.
  \item As for the collaboration of intermediate layers, cross-entropy loss and consistency regularization crossing each layer are leveraged to directly produce local error signal and relieve gradient vanishing problem brought by gradient descent.
  \item We further borrow the idea of manifold learning by measuring the similarity between intermediate feature representations of different layers.
  \item For reducing redundancy in each layer, the collaboration of the convolution kernel is employed as a resultful regularization in training.
\end{itemize}

With a relatively low increase in memory cost, we achieve evident improvements on various datasets.
In addition, selectively combining these ways can yield even better performance.
This work not only solves the problem of huge memory cost of \cite{collaborative} in training stage,
but also explore more possibilities of collaborative learning for different parts in a single network.

Our contributions are summarized as follows:
\begin{enumerate}
  \item We propose a new framework of collaborative learning within one single network. 
  Under this framework, accuracy can be improved with neither extra inference cost nor the 
  enlargement of network capacity.
  \item Compared with \cite{collaborative}, our collaboration framework contains 
  four different ways with lower training memory cost, which is more friendly for 
  memory-limited GPU training case. What's more, our proposed methods are versatile and 
  not limited to the output layer. 
  It's flexible to separately use one of them to yield lower training costs or to jointly use the 
  combination of them to achieve the best performance.
  \item The empirical experiments results on CIFAR-10, CIFAR-100, ImageNet32, STL-10 datasets 
  demonstrate the effectiveness of the proposed methods as well as their combinations. 
  For example, we reduce the test errors on STL-10 by $9.28\%$ and $5.45\%$ with ResNet-18 and VGG-16 
  respectively.
\end{enumerate}

%% file: sections/prior_work.tex
\section{Prior Work}
In \cite{collaborative}, collaborative learning was proposed to simultaneously train several 
heads and learn from the outputs of other heads besides the ground-truth labels. 
In inference stage, only one path is preserved, so that the inference graph is kept
unchanged and no extra inference cost is brought. 
Their collaborative learning framework mainly includes two parts: 
learning objective and patterns of multiple heads.

\paragraph{Learning Objective}
The learning objective contains two losses: $J_{hard}$ loss is the normal cross-entropy loss, 
while $J_{soft}$ loss is the collaboration loss.
To be more specific, for a network with $H$ heads, let $\mathbf{z}^{(h)}=[z_1, z_2, \dots , z_m]^\intercal$ 
denote the prediction logit vector of head $h$, where $m$ is the number of classes and $h$ ranges from $1$ to $H$.
The corresponding softmax with temperature $T$ is given as 
$\psi_i(\mathbf{z}^{(h)}; T) = \exp(z_i^{(h)}/T)/\sum_{j=1}^m \exp(z_j^{(h)}/T)$.
Given the training data $(x, y)$ where $x$ is an input image and $\mathbf{y}=[y_1, y_2, \dots , y_m]$ 
is its target one-hot vector, the $J_{hard}$ loss with temperature $T$ is defined as:
\begin{align}
\label{eq: J_hard_loss}
J_{hard}(\mathbf{y}, \mathbf{z}^{(h)})  =   
-\sum_{h=1}^m  y_i log (\psi_i(\mathbf{z}^{(h)}; 1)) .
\end{align}
To encourage collaboratively learning from the whole population and achieve 
the consensus of multiple views, $J_{soft}$ loss is defined as follows:
\begin{equation}
  \begin{aligned}
    \mathbf{q}^{(h)} &= 
              \psi(\frac{1}{H-1}\sum\limits_{j\neq h}\mathbf{z}^{(j)}; T), \\
    J_{soft}(\mathbf{q}^{(h)}, \mathbf{z}^{(h)}) &= 
              -\sum_{i=1}^m q_i^{(h)} log(\psi_i(\textbf{z}^{(h)}); T). \\
  \end{aligned}
\end{equation}
The final training objective of \cite{collaborative} is written as:
\begin{align}
  \mathcal{L} = \frac{1}{H} \sum_{h=1}^H 
                  \alpha J_{hard}(\mathbf{y}, \mathbf{z}^{(h)}) + 
                  (1-\alpha) J_{soft}(\mathbf{q}^{(h)}, \mathbf{z}^{(h)}) ,
\end{align}
where $\alpha$ is a trade-off parameter (set as $0.5$ in their implementation).

\begin{figure}[!t]
  \centering
  \subfigure[Original network]{
    \begin{minipage}[t]{0.22\textwidth}
    \centering
    \includegraphics[height=1.70in]{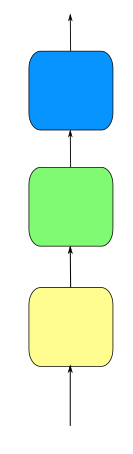}
    \end{minipage}%
    \label{fig: original} 
 }
  \subfigure[Multi-instance]{
     \begin{minipage}[t]{0.22\textwidth}
     \centering
     \includegraphics[height=1.70in]{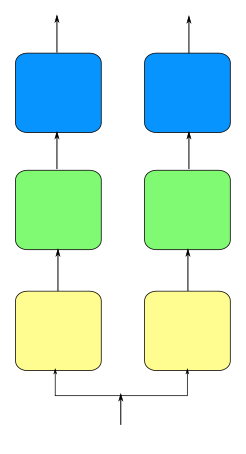}
     \end{minipage}%
     \label{fig: multi_instance}
  }
  \subfigure[Tree-structure]{
     \begin{minipage}[t]{0.22\textwidth}
     \centering
     \includegraphics[height=1.70in]{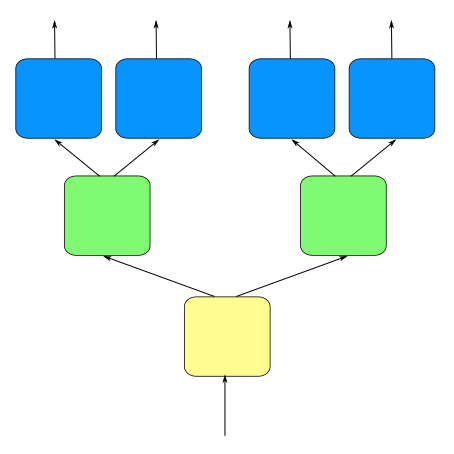}
     \end{minipage}%
     \label{fig: tree_structure}
  } 
  \subfigure[Hierarchical dropout]{
     \begin{minipage}[t]{0.22\textwidth}
     \centering
     \includegraphics[height=1.70in]{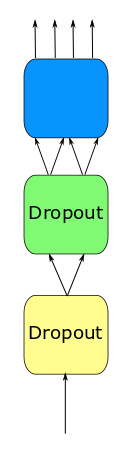}
     \end{minipage}%
     \label{fig: dropout} 
  }
  \caption{Training graphs:
           (a) is original model is composed of three subnets in this illustration.
           (b) and (c) are proposed in \cite{collaborative}. 
           (d) shows our collaborative way of output layer using hierarchical dropout structure.}
  \label{fig: head}
\end{figure}

\paragraph{Patterns of Multiple Head}
Song and Chai \cite{collaborative} proposed two patterns of collaborative learning ---  multi-instance and tree-structure.
Assuming the original network is composed of three subnets, as shown in Fig.\ref{fig: original}. 
Multi-instance, shown in Fig.\ref{fig: multi_instance}, simply duplicates the original network and the memory 
cost is proportional to the number of paths.
Fig.\ref{fig: tree_structure} shows a tree-structure
where intermediate-level representation are 
shared by all subnets in the same stage, thus memory cost is decreased to some extent and generalization is improved.

\paragraph{Connection to other methods}
The collaborative learning method of Song and Chai \cite{collaborative} has originated from previous training algorithms 
by adding additional networks in the training graph to boost accuracy without affecting the inference graph.
To better understand their method and our new framework, we highlight the similarity and some differences between 
their method and other methods here:
\begin{enumerate}
  \item Auxiliary training \cite{aux_training} aims to improve the convergency of network by adding classifiers 
  at specified layers, which will be abandoned in inference stage. 
  By contrast, collaborative learning \cite{collaborative} direct duplicates modules form the original network to avoid the 
  necessity of designing a new structure.
  \item Multi-task training \cite{multi_task, multi_task_2} is proposed to learn multiple related tasks 
  simultaneously so that knowledge learned by different tasks can be reused. 
  The collaborative objective of \cite{collaborative} can be viewed as a special form of multi-task 
  learning where the consensus of multiple views are achieved. Since the collaborative objective is very 
  similar to the original classification objective, there's no need to meticulously design a specified objective 
  for different tasks like the common multi-task training methods. Also collaborative learning can be applicable 
  to single task scenario where multi-task methods cannot help.
  \item Knowledge distillation \cite{distilling} is to train a smaller student model for mimicking the 
  behavior of a larger teacher model in order to achieve model compression and knowledge transfer.
  However, such a teacher model with larger capacity and better performance needs extra work in 
  designing and training. In contrary, collaborative learning doesn't require a pre-trained larger model 
  to assist training; instead, certain consensus of the target network is leveraged for training in a 
  "bootstrap" manner.
\end{enumerate}

\paragraph{Limitations}
We argue that the method of Song and Chai \cite{collaborative} has two limitations. 
First, such multi-instance or tree structure will bring huge memory cost when the number of paths grows,
which may hinder the training of large-capacity models.
Second, their method is limited to output layer while ignoring the intermediate layers, which are also 
essential for training a neural network.
To address limitations, a new framework of collaborative learning is proposed in this paper, with lower memory 
cost in training and more versatile ways for focusing on different parts of a network.

%% file: sections/method.tex
\section{Intra-Model Collaborative Learning}
In this paper, we propose a new framework for collaborative learning, which consists of four 
different objectives to train a single network.

 
\subsection{Collaboration of Output Layer}

The widely used dropout technique \cite{dropout} can split the 
network into different "thinner" sub-networks by temporarily removing some units with 
a specified probability.
Inspired by this, we propose the \textbf{hierarchical dropout structure} 
in a single network rather than building a multi-instance or tree structure with copies of modules like \cite{collaborative}. 
To be specific, as shown in Fig. \ref{fig: dropout}, we sample units at each layer $K$ times by dropout.
As a consequence, some units are dropout and corresponding parameters are omitted so that the network is split into $K$ branches.
In the end, $K^n$ prediction is given for a model of $n$ output layers.
Since different features are handled by different units, the output sequence of such a 
hierarchical dropout structure represents different views, which can be viewed as the 
collaboration of units in the output layer. 
It's clear to see that the proposed structure doesn't increase the capacity of a network 
and only needs to pay a little overhead to store the multiple outputs. 
For example, a ResNet-101 network with input tensor of shape $64\times 3 \times 64 \times 64$ will 
take 8.77GB of memory to train. 
Using tree-structure in Fig. \ref{fig: tree_structure} will take 13.16GB of memory, 
which has exceeded the maximum limit of 11GB memory for GPUs like RTX 2080Ti. 
However, it only takes 8.78GB of memory in training with the proposed hierarchical 
dropout structure with four predictions (same as in Fig.\ref{fig: dropout}),
which is almost equal to the original network.
Besides, since the structure of thr network is kept unchanged in our method, 
it can naturally obviate the issue of unbalanced gradients of different levels in \cite{collaborative}.
Therefore, the back-propagation rescaling trick is not required in our method.

Denote the produced predictions for a classifier of $n$ output layers and $K$ branches as 
$\mathbf{z}=[\mathbf{z}^{(1)}, \mathbf{z}^{(2)}, \dots , \mathbf{z}^{(K^n)}]^\intercal$, 
the collaboration objective for output layer is defined as follows:
\begin{equation}
  \begin{aligned}
    \mathbf{q}^{(i)} =& 
              \psi(\frac{1}{K^n-1}\sum\limits_{j\neq i}\mathbf{z}^{(j)}; T), \\
    \mathcal{L}_{out} =& \frac{1}{K^n}
              \sum_{i=1}^{K^n} 
              \alpha_{out} J_{hard}(\mathbf{y}, \mathbf{z}^{(i)})  \\
              & + (1 - \alpha_{out}) J_{soft}(\mathbf{q}^{(i)}, \mathbf{z}^{(i)})   . \\
  \end{aligned}
\end{equation}
Similar to Eq. \ref{eq: J_hard_loss}, $\psi(\cdot ;T)$ represents for the softmax operation over all classes with temperature $T$.
Experimentally, we set parameters as $K=2$, $T=2$ and $\alpha_{out}=0.5$ for better performance.
The loss $\mathcal{L}_{out}$ will affect all layers through back-propagation.


\subsection{Collaboration of Multiple Intermediate Layers}

We argue that collaborative learning may generalize from output layer to intermediate layers.
In a CNN, an intermediate layer consists of convolution, batch normalization and activation.
In this part, we propose to use a local classifier at each intermediate layer to make prediction
with the intermediate-level representations (feature maps).
Collaboration between the ground-truth one-hot labels and intermediate layers can be achieved 
by measuring the cross-entropy between the local predictions and the targets. 
To be more detailed, a series of local classifiers are built for each intermediate layers, all of 
which is composed of a max pooling layer, a $3\times3$ convolution and a fully connected layer
to obtain the local classification prediction.
Denote the local prediction of the $i$-th intermediate layer for total $m$ classes as 
$\mathbf{z}^{(i)}=[z_1^{(i)}, z_2^{(i)}, \dots , z_m^{(i)}]^\intercal$ and the 
groud-truth one-hot target as $y$, the local classifier loss is defined as follows:
\begin{align}
  J_{hard}^{mid}(\mathbf{y}, \mathbf{z}^{(i)}) = 
  -\sum_{i=1}^m  y_i log (\psi_i(\mathbf{z}^{(i)}; T)) .
\end{align}

Other works such as Local Error Signal \cite{local_error} and HSIC Bottleneck \cite{hsic} also try 
to build a direct connection between intermediate layers and target labels. 
However, they ignore the correlation between different layers and simply set the same objective for all layers. 
To address this, we further propose the $J_{soft}^{mid}$ loss in order to transfer the knowledge of 
high-level layers to low-level layers, which is defined as follows:
\begin{equation}
  \begin{aligned}
    \mathbf{q}^{(i)} &= 
              \psi(\frac{1}{N-i}\sum_{j=i}^N \mathbf{z}^{(j)}; T), \\
    J_{soft}^{mid}(\mathbf{q}^{(i)}, \mathbf{z}^{(i)}) &= 
              -\sum_{k=1}^m q_k^{(i)} log(\psi_k(\textbf{z}^{(i)}); T), \\
  \end{aligned}
\end{equation}
where $N$ represents the total number of layers in a network.
Combining both of them, the final objective for the $i$-th intermediate layer is defined as:
\begin{align}
  \mathcal{L}_{mid}^{(i)} = 
    \alpha_{mid} J_{hard}^{mid}(\mathbf{y}, \mathbf{z}^{(i)}) + 
    \beta_{mid} J_{soft}^{mid}(\mathbf{q}^{(i)}, \mathbf{z}^{(i)})  .
\end{align}
In experiments, we set $\alpha_{mid}=0.05$, $\beta_{mid}=0.05$ and $T=2$. 
Different from collaboration of output layer, 
here the loss $\mathcal{L}_{mid}^{(i)} $ will only update the connection weights of the $i$-th layer 
by local back-propagation.


\subsection{Collaboration of Intermediate Representation with Inputs and Targets}

The above collaborations mainly focus on harnessing consensus among multiple intermediate layers.
It is possible to leverage the relationship between an intermediate layer and inputs or targets 
along the two ends of the spectrum.
To be more specific, we use metric $S(.)$ to measure the similarity of all data points in a mini-batch. 
Given a data sequence with $n$ data points $\mathbf{d}=[d_1, d_2, \dots , d_n]$ in a mini-batch, 
where $d_i \in \mathbb{R}^{C\times W\times H}$ with image width $W$ and height $H$ of $C$ channels.
However, it's non-trivial to model the similarity over such a huge space. 
To address this, we use standard deviation of each feature map as their low-dimension representations,
hence obtain $\mathbf{z}=[z_1, z_2, \dots , z_n]$, where 
$z_i=[z_{i,1}, z_{i,2}, \dots, z_{i,c}]$ and $z_{i, c}=\sigma(d_{i, c}[\dots][\dots])$. 
Here $d_{i, c}$ is the $c$-th channel of  $i$-th data point.
The similarity matrix $S(x)$ of size $n \times n$ is defined for the mean-centered vectors 
$\mathbf{\widetilde{z}}=[\widetilde{z_1}, \widetilde{z_2}, \dots , \widetilde{z_n}]$ obtanined 
by substracting mean from $\mathbf{z}$.
The element in the $i$-th row and $j$-th column of the similarity matrix describes the similarity 
between the $i$-th data points and $j$-th data points, which is measured by cosine similarity metric:
\begin{align}
  s_{ij} = \frac
            { { \widetilde{z_i} }^\intercal  {\widetilde{z_j}}}
            { \left \| \widetilde{z_i} \right \|_2  \left \| \widetilde{z_j} \right \|_2  } .
\end{align}

Datapoints with the same labels are expected to have similar intermediate-level representations, while 
different labels lead to diverse representations. 
Hence, the similarity matrix of intermediate representations is supposed to minimize the distance to 
the target. 
On the contrary, the intermediate representations can be viewed as extracted features that should 
exhibit discrepancy from the input representations.
The final objective for $i$-th layer is named as $\mathcal{L}_{pull-push}^{(i)}$, 
which represents pulling the intermediate representation to targets and pushing it away from 
the inputs:
\begin{equation}
\begin{aligned}
  \mathcal{L}_{pull-push}^{(i)} = 
    & \alpha_{pull} \left \| S(g_i(h^{(i)})) - S(\mathbf{y})  \right \|_F \\
    & - \alpha_{push} \left \| S(g_i(h^{(i)})) - S(\mathbf{x})  \right  \|_F ,
\end{aligned}
\end{equation}
where $\mathbf{x}$ and $\mathbf{y}$ represent the input data points and target one-hot vectors respectively.
We denote $g_i$ the projection operation for $i$-th intermediate layer, where a single convolution is actually used.
In experiments, we linearly increase $\alpha_{pull}$ and decrease  $\alpha_{push}$ through all the layers from 
inputs to outputs. 
Similarly, $\mathcal{L}_{pull-push}$ will only update the connection weights of the $i$-th layer.


\subsection{Collaboration Inside One Convolution Layer}
The above proposed three forms of collaboration are built based on different parts of network or external 
information (i.e., inputs and target labels), while here we study the collaboration among units
inside one single convolution layer. 
In training deep neural networks, co-adaption often occurs when two or 
more hidden units become highly coupling and relies on each other, which not only yields redundant 
information in the network but also brings serious overfitting problem. 
To cope with this issue, Tompson et al. \cite{SpatialDropout} and Ghiasi et al. \cite{DropBlock} proposed to use 2D Dropout 
which is similar to dropout but is applied on the feature map. 
However, the probability of dropout is difficult to 
control and may bring under-fitting. Cogswell et al. \cite{Cogswell2015Reducing} proposed to minimize the 
cross-covariance of hidden activations, but it leads to large computation costs when the size 
of feature map is huge. 

Here a new collaboration method for the weights of convolution kernel $W$ is proposed.
Since restricting feature maps may take huge computation cost, we choose to direct 
regularize $W$ by minimizing its covariance matrix.
This can be seen as a collaboration of convolution kernel weights to reduce the redundancy of the feature maps.
Given the kernel weight $W \in \mathbb R^{F \times C \times W \times H}$ of the specified 
convolution layer, where $F$ is the number of filters, $C$ is the number of channels, 
$W$ and $H$ are the size of kernel weight. 
We first reshape $W$ to a matrix form $W^M \in \mathbb R^{F \times G}$, 
where $G=C \times W \times H$. Zero-mean normalization is used to obtain the normalized 
matrix $\widetilde{W}^M$, in which the $i$-th row is represented by
$\widetilde{W}^M_i = (W^M_i - \mu({W^M_i})/\sigma({W^M_i}) $ with $\mu$ and $\sigma$ represent
for mean and standard deviation respectively. 
Elements in the covariance matrix of the normalized $\widetilde{W}^M$ and the collaboration objective 
are defined as:
\begin{equation}
  \begin{aligned}
    C_{i, j} &= \frac{1}{G}
                \sum_{k=1}^{G} \widetilde{W}^M_{i, k} \widetilde{W}^M_{j, k} , \\
    \mathcal{L}_{kernel} &=  \left \| C - diag(C)  \right \|_F . \\
  \end{aligned}
\end{equation}

The intuition is that the redundancy of feature maps will be reduced if the covariance is controlled for convolution kernels.
In experiments, we found it beneficial to use $\mathcal{L}_{kernel}$ only in the last two groups of convolutions for VGG-16 and ResNet-18.
Since we direct compute loss over convolution kernel, the loss $\mathcal{L}_{kernel}$ only updates the connection weights of the current convolution layer.

%% file: sections/experiments.tex
\newcommand{\tabincell}[2]{\begin{tabular}{@{}#1@{}}#2\end{tabular}}  
\renewcommand{\arraystretch}{1.3} 
\setlength{\tabcolsep}{3.5pt} 

\begin{table*}[!t]{
  \begin{center}
  \caption{\textbf{Test error}($\mathbf{\%}$) on CIFAR-10 and CIFAR-100.}
  \label{table: cifar}
  \scalebox{1.0}{
  \begin{tabular}{l | l | c c | c c c c | c c } 
  \hline
  Dataset & Model & baseline & \tabincell{c}{baseline\\(2x)} & $\mathcal{L}_{out}$ & $\mathcal{L}_{mid}$ & 
  $\mathcal{L}_{pull-push}$ &  $\mathcal{L}_{kernel}$ &
  \small{\tabincell{c}{$\mathcal{L}_{out}$ + $\mathcal{L}_{mid}$ \\ + $\mathcal{L}_{pull-push}$}}  &
  \small{\tabincell{c}{$\mathcal{L}_{pull-push}$ \\ + $\mathcal{L}_{mid}$ + $\mathcal{L}_{kernel}$}}  \\
  \hline 
  \multirow{2}{*}{\small{CIFAR-10}}  & VGG-16    & 6.32 & 5.48 & 5.90 & \textbf{5.44} & 5.51 & 6.04 & \textbf{5.34} & 5.42  \\
  \cline{2-10}
                                     & ResNet-18 & 4.84 & 4.53 & 4.55 & 4.47 & \textbf{4.45} & 4.69 & 4.42 & \textbf{4.31} \\
  \hline
  \multirow{2}{*}{\small{CIFAR-100}} & VGG-16    & 26.94 & 25.39 & 25.90 & \textbf{24.93} & 26.03 & 25.64 & \textbf{24.18} & 25.13  \\
  \cline{2-10}
                                     & ResNet-18 & 22.98 & 21.58 & 22.02 & \textbf{21.93} & 22.27 & 22.55 & 22.03 & \textbf{20.93} \\
  \hline
  \end{tabular}
  }
  \end{center}
}
\end{table*}

\renewcommand{\arraystretch}{1.3} 
\begin{table*}[!t]{
\begin{center}
\caption{\textbf{Top-5 test error}($\mathbf{\%}$) on ImageNet32 and \textbf{top-1 test error}($\mathbf{\%}$) on STL-10.}
\label{table: imagenet32_stl10}
\scalebox{1.0}{
\begin{tabular}{l | l| c c  | c  c c c |c  c } 
\hline
Dataset & Model & baseline &\tabincell{c}{baseline\\(2x)}& $\mathcal{L}_{out}$ & $\mathcal{L}_{mid}$ & 
$\mathcal{L}_{pull-push}$ &  $\mathcal{L}_{kernel}$ &
\tabincell{c}{$\mathcal{L}_{out}$ + $\mathcal{L}_{mid}$ \\ + $\mathcal{L}_{pull-push}$}  &
\tabincell{c}{$\mathcal{L}_{pull-push}$ \\ + $\mathcal{L}_{mid}$  + $\mathcal{L}_{kernel}$}  \\
\hline 
\multirow{2}{*}{ImageNet32} &  VGG-16    & 33.08  & 30.06 & 30.00 & \textbf{29.09} & 29.51 & 31.52 & \textbf{28.41} & 28.75  \\
\cline{2-10}
                            &  ResNet-18 & 24.10  & 22.38 & 22.27 & \textbf{22.01} & 23.12 & 22.95 & 22.05 & \textbf{21.64} \\
\hline
\multirow{2}{*}{STL-10}     &  VGG-16    & 30.25  & 39.70 & 26.35 & 28.87 & \textbf{25.92} & 29.53 & \textbf{24.80} & 25.47  \\
\cline{2-10}
                            &  ResNet-18 & 33.31  & 32.05 & 27.89 & 30.59 & \textbf{24.80} & 30.35 & 24.66 & \textbf{24.03} \\
\hline
\end{tabular}
}
\end{center}
}
\end{table*}

\begin{table*}[!t]{
  \begin{center}
  \caption{\textbf{Test error}($\mathbf{\%}$) under different combination of collaborations on CIFAR-100.}
  \label{table: combine}
  \scalebox{1.0}{
  \begin{tabular}{l  || c| c| c| c|| c |c |c |c |c |c ||c |c |c |c ||c } 
  \hline
  $\mathcal{L}_{out}$        & \checkmark &            &            &            & \checkmark & \checkmark & \checkmark &            &            &            & \checkmark & \checkmark & \checkmark &            & \checkmark \\
  $\mathcal{L}_{mid}$        &            & \checkmark &            &            & \checkmark &            &            & \checkmark & \checkmark &            & \checkmark & \checkmark &            & \checkmark & \checkmark \\
  $\mathcal{L}_{pull-push}$  &            &            & \checkmark &            &            & \checkmark &            & \checkmark &            & \checkmark & \checkmark &            & \checkmark & \checkmark & \checkmark \\
  $\mathcal{L}_{kernel}$     &            &            &            & \checkmark &            &            & \checkmark &            & \checkmark & \checkmark &            & \checkmark & \checkmark & \checkmark & \checkmark \\
  \hline
  VGG-16                     & 25.90      & 24.93      & 26.03      & 25.64      & 24.76      & 25.31      & 25.62      & 24.67      & 24.81      & 25.68      & \textbf{24.18} & 24.39  & 24.81      & 25.13      &  24.89 \\
  ResNet-18                  & 22.02      & 21.93      & 22.27      & 22.55      & 21.79      & 21.84      & 21.88      & 21.60      & 21.81      & 21.86      & 22.03      & 21.42      & 21.89      & \textbf{20.93} &  21.51 \\ 
  \hline
  \end{tabular}
  }
  \end{center}
}
\end{table*}

\section{Experiments}

We first conduct experiments by using each of the proposed collaboration separately to demonstrate 
their effectiveness. In order to achieve a higher accuracy, we further study the combinations 
of the proposed four ways and investigate the relationship between them. 
The results of the best combination on each datasets are also reported.
We use the popular VGG-like (VGG-16) and ResNet-like (ResNet-18) models as backbone networks, followed by 
three fully-connected layers for image classification.
Dropout with a probability of $0.5$ is used in the first two layers of classifier to reduce overfitting.
SGD optimizer with $momentum=0.9$ is used in all experiments, with different learning rate and 
weight decay on different datasets. The experiments are implemented on PyTorch framework, 
using a single RTX 2080Ti GPU with 11GB memory. 

\subsection{Results on Four Datasets}
We first report results of experiments on CIFAR-10, CIFAR-100, ImageNet32 and STL-10 datasets to testify the 
effectiveness of the proposed four ways separately and the best combination. 
Besides, the results on CIFAR-100 with noisy labels, following \cite{collaborative},
are also reported to attest the robustness of proposed methods.
Due to the training time budget, we performed only a single run and report the lowest test 
error in all epochs.

\paragraph{CIFAR-10 and CIFAR-100}
As proposed by Krizhevsky and Hinton \cite{cifar}, CIFAR-10 and CIFAR-100 consist of 50,000 RGB images 
with $32 \times 32$ pixels for training and 10,000 for validating, having 10 and 
100 classes respectively. Following \cite{devries2017improved}, we train models for 
200 epochs. Batch size is set as 128 and weight decay is set as 5e-4.
The initial learning rate is 0.1 and decays by a factor of 0.2 after 60, 120 and 160 epochs.

The results on CIFAR dataset are shown in Table \ref{table: cifar}, 
where baseline(2x) means the number of convolution filters is multiplied by a factor of 2. 
It's clear that all of these four Collaborations gives evident improvement of test accuracy. 
In addition, $\mathcal{L}_{mid}$ has the greatest impact such that it offers competitive accuracy approaching 
to baseline with 2 times of the number of filters. 
For example, by separately using one way of collaborations, 
we obtain test error $\mathbf{21.93\%}$ and $\mathbf{4.47\%}$ on CIFAR-100 and CIFAR-10 using ResNet-18 without 
bringing any extra cost in inference stage. 
What's more, we further reduce the test error to $\mathbf{20.93\%}$ and $\mathbf{4.31\%}$ by 
selectively using the best combination of collaborations.

\paragraph{ImageNet32}
The origin ImageNet dataset \cite{imagenet} is a large-scale classification dataset 
consisting of 1000 object classes. For each class, it contains 50 test samples and more than 
1000 training samples.
Due to the large amount and the relatively large size of images, it tends to take several days to train 
a model on a single GPU. 
Chrabaszcz et al.\cite{chrabaszcz2017downsampled} proposed a downsampled version of ImageNet by downsampling each image to 
a $32\times 32$ size and keeping the amount of images unchanged.
Chrabaszcz et al.\cite{chrabaszcz2017downsampled} proposed a downsampled version of ImageNet. 
Following \cite{chrabaszcz2017downsampled}, we train models for 40 epochs.
Batch size is set as 256 and initial learning rate is set as 0.1, decayed after 12, 24, 36 epochs by a factor of 0.2.

We report the top-5 error in Table \ref{table: imagenet32_stl10}. 
It's clear that our methods can offer evident improvements on this challenging dataset. 
By separately using each way of collaboration, we improve the top-5 accuracy by $\mathbf{3.99\%}$ 
and $\mathbf{2.19\%}$ for VGG-16 and ResNet-18 respectively. 
The combination of collaborations further produces error improvements of $\mathbf{4.67\%}$ 
and $\mathbf{2.46\%}$.

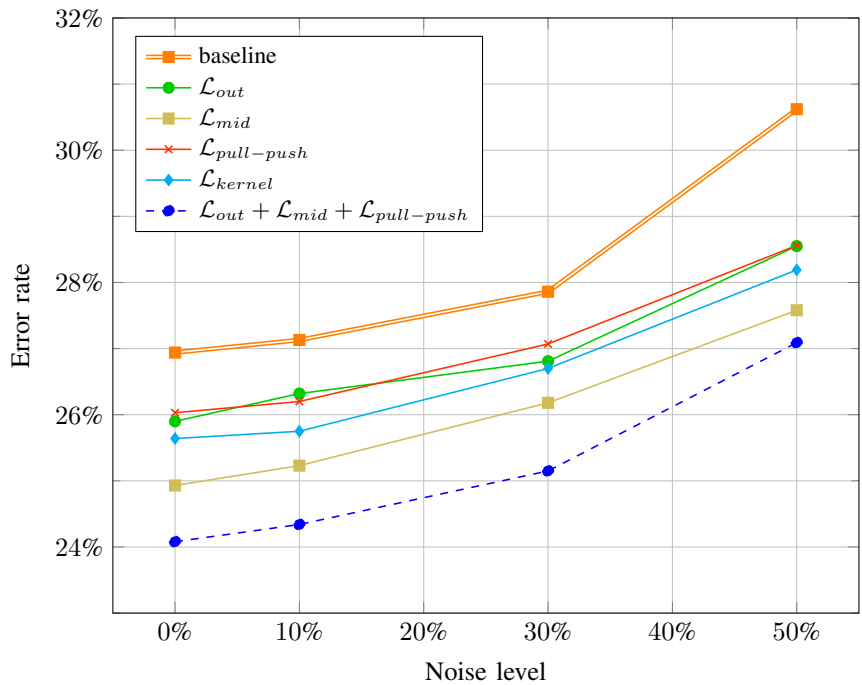
\begin{figure*}[!t]
  \begin{center}
  \begin{tikzpicture}

    \begin{axis}[
      grid=both,
    grid style={line width=.1pt, draw=gray!10},
    major grid style={line width=.2pt,draw=gray!50},
    minor grid style={line width=.2pt,draw=gray!50},
    height=9.5cm,
    width=11.5cm,
    xlabel={Noise level},
    ylabel={Error rate},
    ymin=23, 
    ymax=32,
    xtick={0,10,20,30,40,50},
    legend style={
      cells={anchor=west},
      font=\small,
    },
    xticklabel={\pgfmathprintnumber\tick\%},
    yticklabel={\pgfmathprintnumber\tick\%},
    minor y tick num=1,
    legend pos=north west
    ]

    \addplot[double, color=orange,mark=square*, semithick] coordinates {
      (0, 26.94)
      (10, 27.13)
      (30, 27.86)
      (50, 30.62) 
    };
    \addlegendentry{baseline}
    \addplot[color=green!80!black,mark=otimes*, semithick] coordinates {
      (0.0, 25.90)
      (10, 26.32)
      (30, 26.81)
      (50, 28.55)
    };
    \addlegendentry{$\mathcal{L}_{out}$}
  
    \addplot[color=yellow!80!blue,mark=square*, semithick] coordinates {
      (0.0, 24.93)
      (10, 25.23)
      (30, 26.18)
      (50, 27.58)
    };
    \addlegendentry{$\mathcal{L}_{mid}$}
  
    \addplot[color=red!80!yellow,mark=x, semithick] coordinates {
      (0.0, 26.03)
      (10, 26.20)
      (30, 27.07)
      (50, 28.56)
    };
    \addlegendentry{$\mathcal{L}_{pull-push}$}
  
    \addplot[color=cyan,mark=diamond*, semithick]  coordinates {
      (0, 25.64)
      (10, 25.75)
      (30, 26.70)
      (50, 28.19)
    };
    \addlegendentry{$\mathcal{L}_{kernel}$}
  
    \addplot[dashed, color=blue ,mark=otimes*, semithick] coordinates {
      (00, 24.08)
      (10, 24.34)
      (30, 25.15)
      (50, 27.09)
    };
    \addlegendentry{$\mathcal{L}_{out}+\mathcal{L}_{mid}+\mathcal{L}_{pull-push}$}
    \end{axis}
    \end{tikzpicture}
  \end{center} 
  \caption{Test error of VGG-16 on CIFAR-100 with label noise. Noise level is the portion 
           of labels which are uniformly sampled from the whole class labels.}
  \label{fig: noise_label}
\end{figure*}

\paragraph{STL-10}
STL-10 \cite{coates2011analysis} is a classification dataset with 10 object classes. There're 
500 training images and 800 test images per class with each image of $96\times 96$ pixels.
We train models for 200 epochs, where batch size is set as 128 and weight decay is set as 5e-4.
The initial learning rate is set as 0.05 and decayed after 60, 120, 180 epochs by a factor of 0.2.
Note that for ResNet-18 with 2x number of filters, we decrease the batch size by half due to memory 
limitation.

As shown in Table \ref{table: imagenet32_stl10}, $\mathcal{L}_{pull-push}$ yields significant
improvements up to $\mathbf{8.51\%}$ for ResNet-18 and $\mathbf{4.33\%}$ for VGG-16. 
By using $\mathcal{L}_{out}$ or $\mathcal{L}_{mid}$, we can also gain remarkable improvements. 
However, $\mathcal{L}_{kernel}$ seems to be not that effective for VGG-16. 
By jointly using multiple collaborations, we yield remarkable accuracy improvements up to 
$\mathbf{5.45\%}$ and $\mathbf{9.28\%}$ for VGG-16 and ResNet-18 respectively. 

\paragraph{Robustness to label noise}
Following \cite{collaborative}, we conduct experiments on CIFAR-100 with VGG-16 to validate the 
noisy label resistance of the proposed methods. 
Noisy labels are corrupted with a uniform distribution over the whole labels. 
The portion of noisy labels are fixed in a single run, but noisy labels are randomly 
generated every epoch. As shown in Fig. \ref{fig: noise_label}, we execute experiments over noise levels 
range from $10\%$ to $50\%$. It's evident that the above four methods of collaborative learning as well 
as their combination yield significant improvements compared to baseline. 
What's more, the accuracy gains become larger when the portion of noisy labels is huge. 
For example, $\mathcal{L}_{out}+\mathcal{L}_{mid}+\mathcal{L}_{pull-push}$ offers an improvement of 
$\mathbf{3.53\%}$ over baseline at the noise level of $50\%$, which demonstrates that our methods are more tolerant
to noisy labels.

\subsection{Combination of Collaborations}  
\label{section: combination} 

Since the effectiveness of using collaboration separately has been demonstrated, 
we further conduct experiments to investigate the effectiveness of different combinations of collaborations. 
Due to time limits, experiments with VGG-16 and ResNet-18 on CIFAR-100 are reported in Table \ref{table: combine}.

The optimal weights for each form of collaboration are obtained by grid-search approach and 
it's clear that that the best performance cannot be achieved by simply stacking all of these ways of collaboration.
For VGG-16, $\mathcal{L}_{out}$ tends to offer improvements when combined with others. 
Using both $\mathcal{L}_{pull-push}$ and $\mathcal{L}_{mid}$ can also bring evident improvements, while further 
using $\mathcal{L}_{kernel}$ increase the test error instead.
As is shown, using $\mathcal{L}_{out}$ + $\mathcal{L}_{mid}$ + $\mathcal{L}_{pull-push}$ yields the lowest test error.
For ResNet-18, we found it most beneficial to use  $\mathcal{L}_{mid}$ + $\mathcal{L}_{pull-push}$ +  $\mathcal{L}_{kernel}$.
With the explored combinations, for example, we further reduce the test errors by $\mathbf{0.16\%}$, $\mathbf{1.00\%}$,
$\mathbf{0.37\%}$ and $\mathbf{0.77\%}$ compared to separately using one best form of collaborations on CIFAR-10, CIFAR-100, ImageNet32 and 
STL-10 respectively.

%% file: sections/conclusion.tex
\section{Conclusion}
In this paper, we propose a new framework of collaborative learning with the following features:
\begin{enumerate}
  \item Although the method proposed by Song and Chai \cite{collaborative} doesn't require extra inference cost, 
  the memory overhead might be huge for training large-capacity models on memory limited GPUs.
  In contrast, our proposed framework significantly lessens the memory burden in training.
  \item Compared to the method of Song and Chai\cite{collaborative} which totally depends on multi-head patterns to 
  achieve consensus, our new four ways of collaboration offer versatile consensus among different 
  part of a single network and provide higher flexibility for single deployment or selective combination.
  \item Results on four datasets testify the improvements brought by each of these four methods. 
  Besides, the best combinations can be found to offer further improvements in accuracy.
  \item We demonstrate that the proposed methods can still yield better performance under relatively 
  high levels of noisy labels, which verifies the robustness of our framework.
\end{enumerate}

In the future, we are planning to extend collaborative learning to other fields such as semantic segmentation,
object detection and person re-identification.